\documentclass[runningheads]{llncs}
\usepackage[T1]{fontenc}
\usepackage{graphicx}
\usepackage{booktabs}
\usepackage{multirow}
\usepackage{amsmath}
\usepackage{amssymb}
\usepackage{pifont}
\usepackage[table]{xcolor}
\usepackage{booktabs}
\usepackage{siunitx}
\newcommand{\cmark}{${\checkmark}$}
\newcommand{\xmark}{\textcolor{gray!50}{${\times}$}}
\usepackage{makecell}
\usepackage{bbding}

\begin{document}
\title{Towards Unified Surgical Scene Understanding: Bridging Reasoning and Grounding via MLLMs}
\author{
Jincai Huang\inst{1}\textsuperscript{\textdagger} \and
Shihao Zou\inst{2}\textsuperscript{\textdagger} \and
Yuchen Guo\inst{3} \and
Jingjing Li\inst{4} \and
Wei Ji\inst{5} \and
Kai Wang\inst{6} \and
Shanshan Wang\inst{2} \and
Weixin Si\inst{7}\Envelope
}
\authorrunning{J. Huang et al.}
\institute{
Southern University of Science and Technology
\and
Shenzhen Institutes of Advanced Technology, Chinese Academy of Sciences
\and
Northwestern University
\and
University of Alberta
\and
Yale University
\and
Nanfang Hospital
\and
Shenzhen University of Advanced Technology\\
\email{siweixin@suat-sz.edu.cn}\\
\textdagger\ These authors contributed equally.
}
\maketitle
\begin{abstract}
Surgical scene understanding is a cornerstone of computer-assisted intervention. While recent advances, particularly in surgical image segmentation, have driven progress, real-world clinical applications require a more holistic understanding that jointly captures procedural context, semantic reasoning, and precise visual grounding. However, existing approaches typically address these components in isolation, leading to fragmented representations and limited semantic consistency. To address this limitation, we propose SurgMLLM, a unified surgical scene understanding framework that bridges high-level reasoning and low-level visual grounding within a single model. Given surgical videos, SurgMLLM fine-tunes a multi-modal large language model (MLLM) to support structured interpretability reasoning, which is used to jointly model phases, instrument-verb-target ($IVT$) triplets, and triplet-entity segmentation tokens. These tokens are then temporally aggregated and serve as prompts for a segmentation network, enabling accurate pixel-wise grounding of triplet instruments and targets. The entire framework is trained end-to-end with a unified objective that couples language-based reasoning supervision with visual grounding losses, promoting coherent cross-task learning and clinically consistent scene representations. To facilitate unified evaluation, we introduce CholecT45-Scene, extending CholecT45 dataset with 64,299 frames of pixel-level mask annotations for instruments and targets, aligned with existing triplet labels. Extensive experiments show that SurgMLLM significantly advances surgical scene understanding, improving the primary triplet recognition metric AP$_{IVT}$ from 40.7\% to 46.0\% and consistently outperforming prior methods in phase recognition and segmentation. These results highlight the effectiveness of unified reasoning-and-grounding for reliable, context-aware surgical assistance. The code and dataset will be released.
\end{abstract}
\section{Introduction}
Minimally invasive and robot-assisted surgery~\cite{long2025surgical} relies primarily on endoscopic videos as the main source of intra-operative information. To enable context-aware computer-assisted intervention, an intelligent system should not only recognize workflow progression over time, but also reason about fine-grained instrument--tissue interactions and ground these semantics to image regions for interpretable decision support~\cite{pan2026surgical}. However, achieving coherent reasoning across temporal, relational, and spatial dimensions remains a fundamental challenge.
\begin{figure}[t]
    \includegraphics[width=\textwidth]{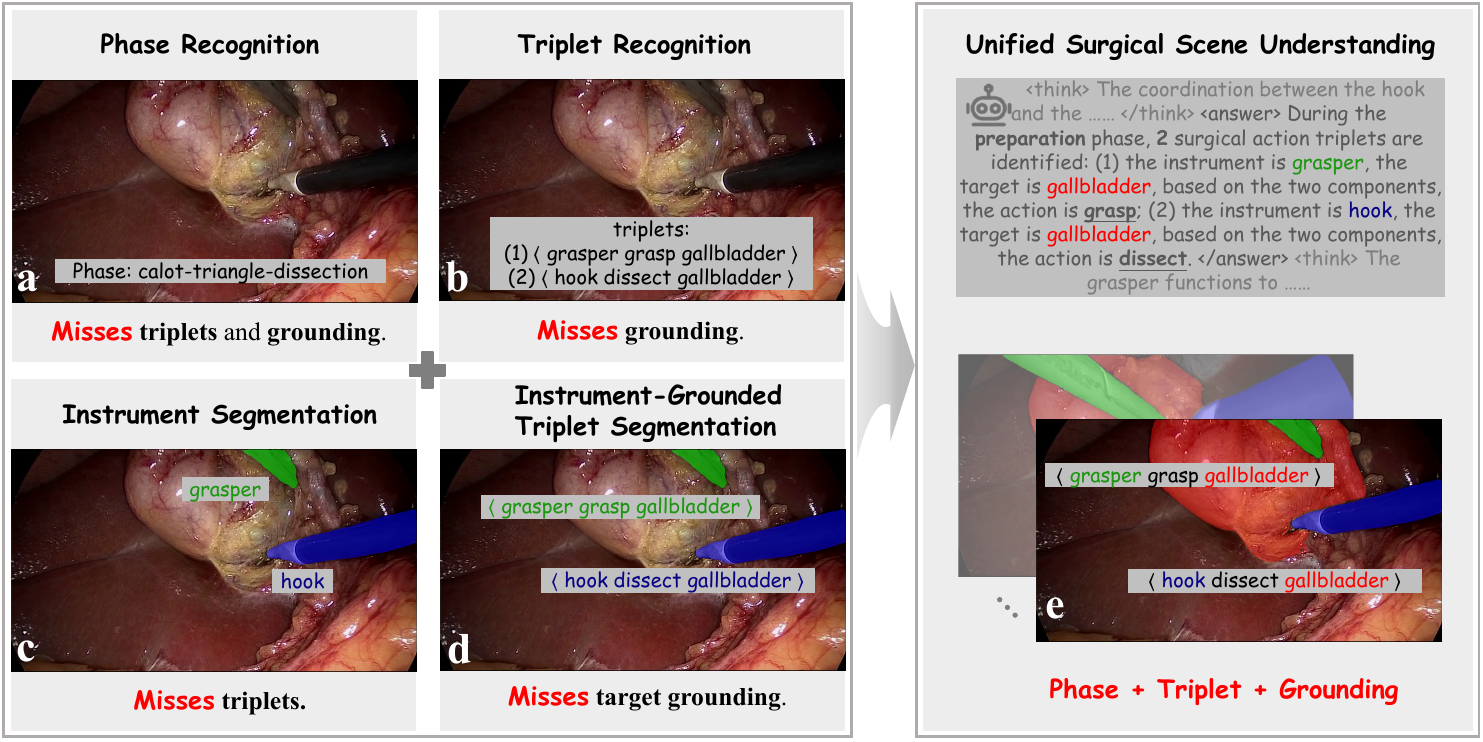}
    \vspace{-6mm}
    \caption{\textbf{Representative tasks in surgical scene understanding.} (a) Phase recognition~\cite{lan2025new} provides only high-level workflow state prediction. (b) Triplet recognition~\cite{jeon2025curconmix} identifies only high-level $IVT$ interactions without visual grounding. (c) Instrument segmentation~\cite{li2023structural} performs only pixel-wise semantic recognition of surgical instruments. (d) Recent instrument-grounded triplet segmentation~\cite{alabi2025grounding} predicts triplet entities while segmenting instruments at the pixel level, but neglecting phase recognition and target grounding. (e) Our framework bridges high-level phase and triplet reasoning with low-level triplet entity grounding via MLLM, enabling comprehensive and coherent surgical scene understanding for robust computer-assisted intervention.}
\label{fig:overview}
\end{figure}
Prior research has advanced key components of surgical scene understanding, yet largely in isolation. For temporal workflow modeling, phase recognition evolved from convolution-recurrent pipelines to transformer-based long-range modeling (\textit{e.g.}, Trans-SVNet~\cite{gao2021trans} and SurgFormer~\cite{lan2025new}). Complementarily, triplet recognition models $IVT$ interactions~\cite{nwoye2022rendezvous,gui2023mt4mtl,jeon2025curconmix}, while instrument segmentation~\cite{liusurgical,ceron2022real} provides pixel-level localization without interaction reasoning. Recent instrument-grounded triplet segmentation methods~\cite{sharma2023surgical,alabi2025grounding} begin to couple prediction and grounding, but often treat targets as categorical labels with weak spatial constraints, and typically omit phase recognition or full triplet-entity grounding. As shown in Fig.~\ref{fig:overview}, this fragmented modeling hinders coherent understanding across temporal, relational, and spatial dimensions.
Importantly, \textit{surgical scenes exhibit strong intrinsic priors}: workflow phases constrain plausible $IVT$ interactions, while recurring $IVT$ patterns provide cues about phase progression. Ignoring these dependencies prevents models from fully leveraging domain knowledge. Moreover, for computer-assisted intervention, high-level predictions alone are insufficient--semantic decisions must be grounded to precise image regions to ensure reliability and verifiability~\cite{carstens2025artificial}. When attempting to unify these tasks, additional challenges arise, including aligning workflow evolution with interaction semantics, enforcing cross triplet-entity consistency, and maintaining temporally coherent pixel-level grounding.
Meanwhile, MLLM-based grounded reasoning has shown promise in bridging language reasoning and pixel-level representations. The SAM family~\cite{kirillov2023segment,ravisam} provides prompt-driven segmentation primitives, and pixel-grounded conversational models~\cite{rasheed2024glamm,liu2023visual,liuunipixel} further integrate segmentation priors into reasoning pipelines. Token-based designs such as LISA~\cite{lai2024lisa} and Sa2VA~\cite{yuan2025sa2va} demonstrate an emerging trend toward reasoning-guided mask prediction. However, these generic frameworks do not explicitly encode structured surgical semantics (\textit{e.g.}, $IVT$ triplets and workflow phases) nor ensure temporally consistent grounding in surgical videos, motivating a unified surgical-specific MLLM framework.
To address these limitations, we propose \textbf{SurgMLLM}, a unified framework that integrates phase recognition, triplet reasoning, and triplet-entity grounding within an MLLM-driven architecture. Given surgical videos, SurgMLLM first performs structured scene reasoning through explicit textual analysis, modeling intrinsic correlations between workflow phases and $IVT$ entities. The generated entity-aware \texttt{[SEG]} tokens serve as language-conditioned prompts that bridge high-level semantic reasoning and low-level pixel decoding via SAM2~\cite{ravisam}. A temporal fusion mechanism further enforces cross-frame consistency for grounded entities. By jointly optimizing reasoning and segmentation objectives, the model aligns what it infers, predicts, and grounds within a coherent pipeline.
Our contributions are summarized as follows: (1) We introduce the first unified MLLM-based framework for comprehensive surgical scene understanding, jointly addressing workflow phase recognition, $IVT$ triplet reasoning, and triplet-entity grounding within a single model. (2) We design an explicit reasoning-to-grounding bridge via dedicated tokens and temporal prompt fusion, enabling semantically aligned and temporally consistent segmentation. (3) We construct the CholecT45-Scene dataset with 64,299 frames of triplet pixel-level annotations and structured reasoning narratives to support unified supervision. (4) Extensive experiments demonstrate consistent improvements across all three tasks, establishing a strong benchmark for integrated surgical scene understanding.
\section{Method}

\begin{figure}[t]
    \includegraphics[width=\textwidth]{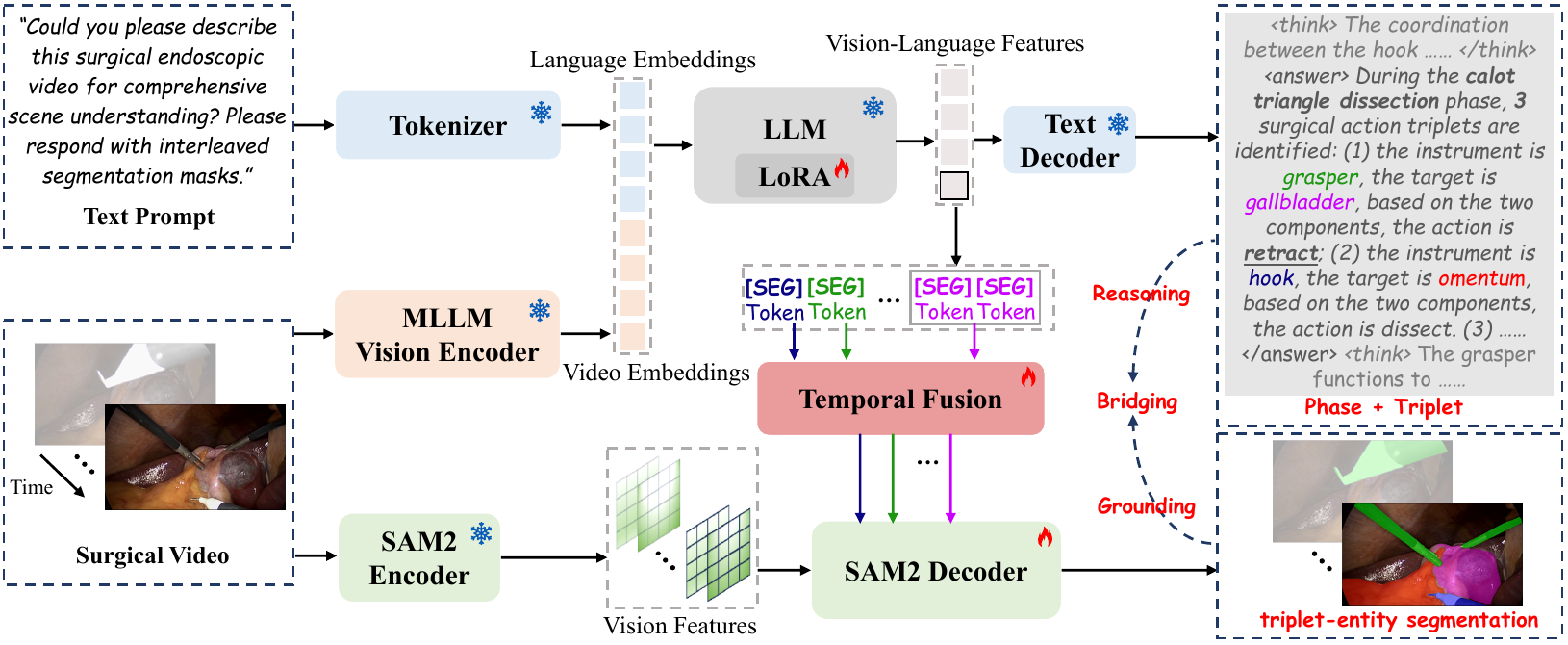}
    \vspace{-6mm}
    \caption{\textbf{Overview of SurgMLLM.} Given surgical videos, SurgMLLM first performs structured scene reasoning with an MLLM to predict workflow phases and $IVT$ triplets, accompanied with triplet-entity \texttt{[SEG]} tokens as language-conditioned prompts. These prompts are temporally fused and inserted into SAM2 to decode pixel-level masks, thereby bridging high-level semantic reasoning and temporally consistent triplet-entity grounding within a unified framework.}
    \label{fig:pipeline}
\end{figure}

\noindent\textbf{Problem Definition.} The overall pipeline is shown in Fig.~\ref{fig:pipeline}. Given a $T$-frame surgical video $\mathcal{X}=\{F_t\}_{t=1}^{T}$ as input, where each frame $F_t\in\mathbb{R}^{H\times W\times 3}$ has height $H$ and width $W$, the goal is to jointly predict temporal semantics and spatial grounding for each frame. Specifically, for the $t$-th frame, the semantics include a workflow phase label $p_t$, a set of surgical triplets $\{(I_{t,n},\ V_{t,n},\ O_{t,n})\}_{n=1}^{N}$, and the corresponding pixel-level grounding masks for the instrument $I_{t,n}$ and target $O_{t,n}$, denoted as $M^{I}_{t,n},\ M^{O}_{t,n} \in \{0,1\}^{H\times W}$. To align with MLLM formulation, both phase labels and triplets are represented as textual categories.
\noindent\textbf{MLLM-Based Reasoning.} Given a textual instruction $q$ and video frames $\mathcal{X}$, we first tokenize $q$ into language tokens $x_q$, and encode $\mathcal{X}$ into visual tokens using a vision encoder followed by a projection to the LLM hidden dimension, producing $x_v$. The concatenated sequence $[x_q, x_v]\in\mathbb{R}^{L\times D^{\text{llm}}}$ is fed into the LLM as a unified multi-modal input. The LLM auto-regressively generates structured textual outputs $x^{\text{llm}}$. The overall reasoning process is formulated as
\begin{equation}
    x^{\text{llm}}=\mathtt{LLM}\big([\mathtt{Tokenize}(q),\ \mathtt{VisionEncoder}(\mathcal{X})]\big).
\end{equation}
Specifically, for the $t$-th frame, the generated output $x^{\text{llm}}_t$ is structured into two components: a \texttt{<think>}...\texttt{</think>} segment that explicitly describes the model's analysis of the surgical scene, followed by a \texttt{<answer>}...\texttt{</answer>} segment that provides the final predictions of the workflow phase and action triplets. 
\begin{align}
    x^{\text{llm}}_t=& \text{``\texttt{<think>}...\texttt{</think>}} \nonumber\\
    &\text{\texttt{<answer>} During } \langle p_t\rangle\text{ phase, } \langle N\rangle\text{ surgical triplet(s) is/are identified:} \nonumber\\
    & \text{(1) instrument is } \langle I_{t,n} \rangle\text{ \texttt{[SEG]}, target is } \langle O_{t,n} \rangle\text{ \texttt{[SEG]}, action is } \langle V_{t,n} \rangle\text{.} \nonumber\\
    & \text{(2)... \texttt{</answer>}''} \nonumber
\end{align}
During fine-tuning, the intrinsic correlations between workflow phases and triplet entities are encoded into the LLM, enabling explicit high-level cross-entity reasoning grounded in the visual context. To further enable spatial grounding, we insert an entity-specific \texttt{[SEG]} token as marker immediately after each triplet-entity token (instrument or target), structured as $\langle I_{t,n}\rangle\ \texttt{[SEG]}$ and $\langle O_{t,n}\rangle\ \texttt{[SEG]}$. These markers function as language-conditioned prompts for downstream segmentation, thereby establishing a direct bridge between high-level semantic reasoning (phase and triplets) and low-level visual grounding.
\noindent\textbf{SAM2-Based Grounding.} Given the structured LLM outputs, we first extract all \texttt{[SEG]} tokens and project them into the SAM2 prompt space, yielding \texttt{[SEG]} prompt embeddings $z\in\mathbb{R}^{N_{\text{seg}}\times D_{\text{sam}}}$, where $N_{\text{seg}}$ denotes the number of \texttt{[SEG]} markers and $D_{\text{sam}}$ is the SAM2 prompt dimension. \textit{Note that for each triplet entity, we assign distinct per-frame \texttt{[SEG]} tokens} rather than using a single shared token across frames. To enforce temporal consistency across per-frame tokens, we group embeddings corresponding to the same entity $u$ (instrument or target) and aggregate them using a residual fusion strategy:
\begin{equation}
    \tilde{z}_n = z_n+ \mathtt{Proj}\Big(\frac{1}{|\mathcal{S}(u)|}\sum_{k\in\mathcal{S}(u)} z_{k}\Big),
\end{equation}
where $\mathcal{S}(u)$ indexes all occurrences of entity $u$ within the input video. This fusion strategy suppresses per-frame noise, yielding more stable and temporally coherent prompts for SAM2-based mask decoding.
Conditioned on the fused prompt $\tilde{z}$, we first encode the video frames using the SAM2 image encoder and subsequently decode the corresponding segmentation masks. The SAM2 features retain fine-grained spatial details, while the \texttt{[SEG]} prompts carry entity-specific semantic cues derived from the LLM reasoning. By combining these complementary representations, the decoded masks are semantically aligned with the predicted triplet instruments and targets on a frame-by-frame basis, enabling consistent and precise grounding.
\noindent\textbf{Losses.} We fine-tune SurgMLLM using the loss function defined as
\begin{equation}
    \mathcal{L}=\mathcal{L}_{\text{llm}} +\lambda_{\text{bce}}\cdot \mathcal{L}_{\text{bce}} +\lambda_{\text{dice}}\cdot\mathcal{L}_{\text{dice}} +\lambda_{\text{ent}}\cdot\mathcal{L}_{\text{ent}}.
\end{equation}
The language modeling loss $\mathcal{L}_{\text{llm}}$ is a standard next-token cross-entropy applied to the structured \texttt{<think>} and \texttt{<answer>} outputs. For spatial grounding, we supervise each \texttt{[SEG]}-conditioned mask using a combination of BCE and Dice losses, $\mathcal{L}_{\text{bce}}$ and $\mathcal{L}_{\text{dice}}$, to ensure accurate and robust pixel-level predictions. Additionally, we introduce an entity-aware loss $\mathcal{L}_{\text{ent}}$ that up-weights token positions corresponding to key semantic elements (phase, instrument, verb, and target), encouraging more precise reasoning over critical surgical entities. 
\begin{figure*}[h]
    \centering
    \begin{minipage}{\textwidth}
        \centering
        \setlength{\tabcolsep}{4pt}
        \resizebox{\textwidth}{!}{
        \begin{tabular}{l c c c c r}
            \toprule
            \rowcolor{gray!20}
            \textbf{Dataset} & \textbf{Phase} & \textbf{Triplet} & 
            \makecell[c]{\textbf{Instrument Mask}} & 
            \makecell[c]{\textbf{Target Mask}} & 
            \textbf{Frames} \\
            \midrule
            CholecT45~\cite{nwoye2023cholectriplet2021} 
                & \cmark & \cmark & \xmark & \xmark & 90,489 \\
            CholecTriplet-Seg~\cite{alabi2025grounding} 
                & \cmark & \cmark & \cmark & \xmark & 30,955 \\
            \midrule
            \textbf{CholecT45-Scene} 
                & \cmark & \cmark & \cmark & \cmark & 64,299 \\
            \bottomrule
        \end{tabular}
        }
    \end{minipage}
    \begin{minipage}{\textwidth}
        \centering
        \includegraphics[width=\textwidth]{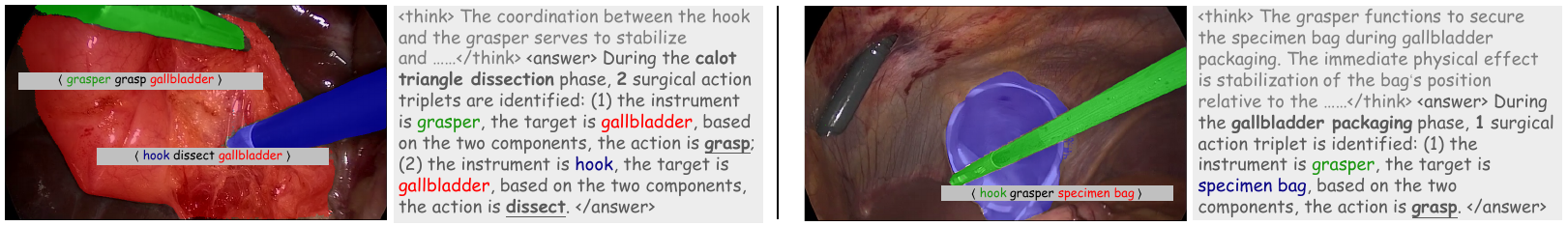}
    \end{minipage}
    \vspace{-4mm}
    \caption{\textbf{CholecT45-Scene Dataset Overview.} Top: Compared with existing surgical datasets, our dataset provides comprehensive annotations including phases, triplets, and triplet-entity masks. Bottom: We visualize annotations in our dataset, presenting instrument and target mask overlays alongside reasoning narratives by Qwen3~\cite{bai2025qwen3}. 
    }
    \label{fig:dataset_overview}
\end{figure*}
\section{Experiments}
\begin{figure}[htb]
    \includegraphics[width=\textwidth]{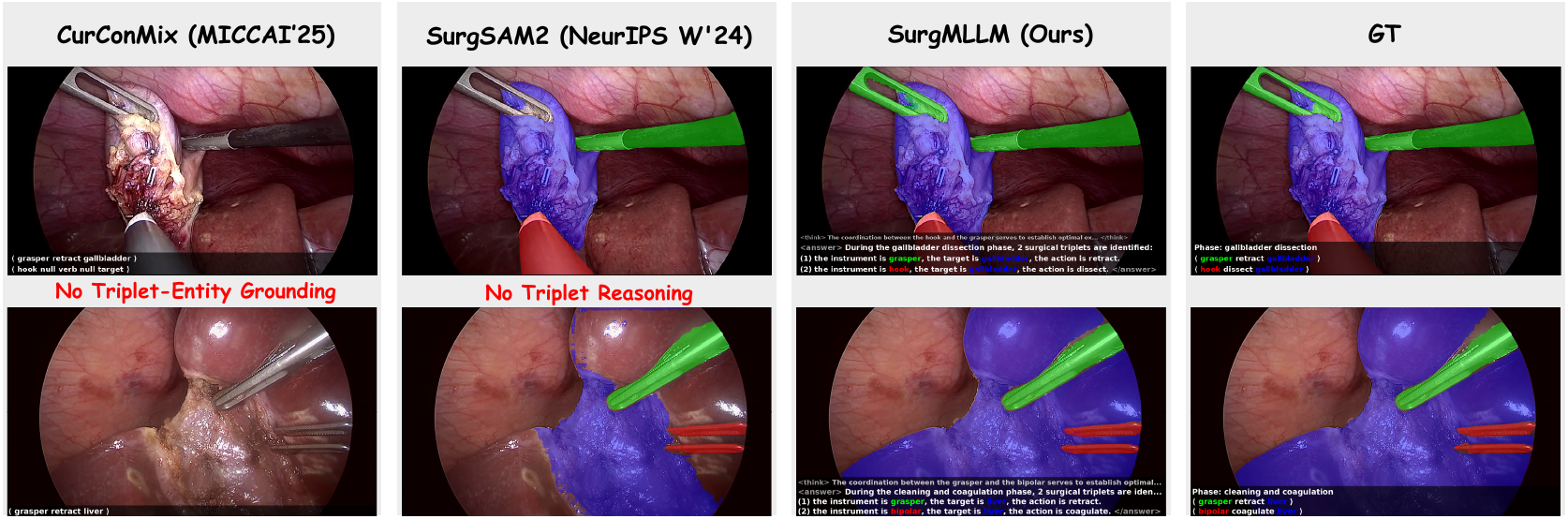}
    \vspace{-4mm}
    \caption{\textbf{Qualitative comparison.} We compare our SurgMLLM with SOTA methods including CurConMix~\cite{jeon2025curconmix} and SurgSAM2~\cite{liusurgical}. Best viewed when zoomed in.}
    \label{fig:qualitative}
\end{figure}

\noindent\textbf{CholecT45-Scene Dataset.} CholecT45~\cite{nwoye2023cholectriplet2021} provides frame-level annotations of workflow phase (7 categories) and $IVT$ triplet entities (6 instruments, 10 verbs, 15 targets). \textit{We extend it with pixel-wise mask annotations for both instruments and corresponding targets.} Specifically, expert surgeons annotate representative key-frames within triplet-consistent video segments using $\sim$30 point prompts to guide SAM2 mask generation, followed by temporal propagation and manual quality auditing to ensure reliability. In addition, we generate structured reasoning narratives for each annotated frame, conditioned on the image, masks, and ground-truth phase and triplet labels, enabling supervision for reasoning-based grounding. The resulting CholecT45-Scene contains 64,299 annotated frames from 45 videos, preserving all original phase and triplet labels while adding dense spatial grounding and reasoning annotations for unified surgical scene understanding (see Fig.~\ref{fig:dataset_overview}).
\noindent\textbf{Evaluation Metrics.} We evaluate three tasks: phase recognition, triplet recognition, and triplet-entity segmentation. For phase recognition, we report video-level Accuracy and phase-level Precision, Recall, and Jaccard under the unrelaxed protocol~\cite{lan2025new}. Triplet recognition is assessed using official Average Precision-based metrics~\cite{nwoye2022rendezvous,jeon2025curconmix}, including component APs ($\text{AP}_I$, $\text{AP}_V$, $\text{AP}_T$), association APs ($\text{AP}_{IV}$, $\text{AP}_{IT}$), and the primary metric $\text{AP}_{IVT}$. For segmentation, we compute IoU for instruments and targets ($\text{IoU}_I$, $\text{IoU}_T$) and report mIoU across entities. Following~\cite{gui2024tail}, we adopt the official 5-fold cross-validation and report mean $\pm$ standard deviation, with ablations conducted on Fold 1.
\noindent\textbf{Implementation Details.} SurgMLLM employs InternVL2.5-4B~\cite{chen2024expanding} as MLLM, with both the vision encoder (InternViT) and LLM frozen, and adapts the model using LoRA (rank 128, $\alpha=256$, dropout 0.05). Input frames are dynamically tiled into $448\times448$ patches. For grounding, we adopt SAM2-H~\cite{ravisam} and fine-tune only the mask decoder on $1024\times1024$ resized frames. LLM hidden states are projected into the SAM2 prompt space via a two-layer MLP, and temporally fused using a lightweight MLP aggregation module. All trainable parameters are optimized with AdamW with learning rate being $6\times10^{-5}$.
\noindent\textbf{Quantitative Results.} Across all three tasks, SurgMLLM consistently achieves new SOTA performance, demonstrating the effectiveness of unified reasoning and grounding. For \textit{phase recognition} (Tab.~\ref{tab:phase_comparison}), it attains the best results on all metrics, notably improving Recall (87.8\%) and Jaccard (69.1\%), indicating enhanced modeling of workflow transitions. For \textit{triplet recognition} (Tab.~\ref{tab:triplet_comparison}), SurgMLLM sets a new benchmark on the primary metric $\text{AP}_{IVT}$ (46.0\% vs. 40.7\%), with especially large gains in target AP and association metrics, reflecting stronger $IVT$ interaction reasoning. The most significant improvements appear in \textit{triplet-entity segmentation} (Tab.~\ref{tab:segmentation_comparison}), where our method markedly outperforms SAM2-based baselines, achieving 87.8\% $\text{IoU}_I$, 67.8\% $\text{IoU}_T$, and 84.4\% mIoU. 
\begin{table}[h]
    \centering
    \caption{Comparison on phase recognition. Best results are in \textbf{bold}.
    }
    \label{tab:phase_comparison}
    \setlength{\tabcolsep}{4pt}
    \resizebox{0.95\textwidth}{!}{
    \begin{tabular}{l c c c c}
        \toprule
            \multirow{2}{*}{\textbf{Method}} & \multicolumn{1}{c}{\textbf{Video-level}} & \multicolumn{3}{c}{\textbf{Phase-level}} \\
        \cmidrule(lr){2-2} \cmidrule(lr){3-5}
            & Accuracy & Precision & Recall & Jaccard \\
        \midrule
            Trans-SVNet (MICCAI'21)~\cite{gao2021trans} & 76.3$\pm$2.4 & 75.1$\pm$3.5 & 74.4$\pm$2.4 & 58.6$\pm$2.2 \\
            SAHC (TMI'22)~\cite{ding2022exploring} & 82.0$\pm$3.7 & 78.1$\pm$1.9 & 76.2$\pm$3.9 & 64.0$\pm$4.1 \\
            SurgFormer (MIA'25)~\cite{lan2025new} & 83.9$\pm$2.3 & 81.1$\pm$2.6 & 77.7$\pm$3.9 & 66.4$\pm$4.2 \\
            \textbf{SurgMLLM (Ours)} & \textbf{84.2$\pm$0.6} & \textbf{86.3$\pm$4.1} & \textbf{87.8$\pm$2.6} & \textbf{69.1$\pm$3.4}  \\
        \bottomrule
    \end{tabular}
    }
\end{table}
\begin{table}[h]
    \centering
    \caption{Comparison on triplet recognition. Best results are in \textbf{bold}.}
    \label{tab:triplet_comparison}
    \setlength{\tabcolsep}{2pt}
    \resizebox{\textwidth}{!}{
    \begin{tabular}{l c c c c c c}
        \toprule
            \textbf{Method} & $\text{AP}_I$ & $\text{AP}_V$ & $\text{AP}_T$ & $\text{AP}_{IV}$ & $\text{AP}_{IT}$ & $\text{AP}_{IVT}$ \\
        \midrule
            RDV (MIA'22)~\cite{nwoye2022rendezvous} & 89.3$\pm$2.1 & 62.0$\pm$1.3 & 40.0$\pm$1.4 & 34.0$\pm$3.3 & 30.8$\pm$2.1 & 29.4$\pm$2.8 \\
            MT4MTL-KD (TMI'23)~\cite{gui2023mt4mtl} & 93.1$\pm$2.1 & 71.8$\pm$3.4 & 48.8$\pm$3.8 & 44.9$\pm$2.4 & 43.1$\pm$2.0 & 37.1$\pm$0.5 \\
            TERL (MICCAI'24)~\cite{gui2024tail} & \textbf{93.5$\pm$2.4} & \textbf{72.8$\pm$2.8} & 51.3$\pm$3.8 & 47.0$\pm$5.6 & 45.7$\pm$2.8 & 38.9$\pm$2.5 \\
            CurConMix (MICCAI'25)~\cite{jeon2025curconmix} & 91.7$\pm$2.2 & 69.5$\pm$0.4 & 51.3$\pm$2.9 & 46.3$\pm$5.0 & 47.1$\pm$1.6 & 40.7$\pm$2.1 \\
            \textbf{SurgMLLM (Ours)} & 89.9$\pm$1.9 & 71.0$\pm$6.1 & \textbf{62.2$\pm$2.1} & \textbf{68.5$\pm$5.8} & \textbf{50.7$\pm$2.8} & \textbf{46.0$\pm$2.8} \\
        \bottomrule
    \end{tabular}
    }
\end{table}
\begin{table}[htb!]
    \centering
    \caption{Comparison on triplet-entity grounding. Best results are in \textbf{bold}.}
    \label{tab:segmentation_comparison}
    \setlength{\tabcolsep}{4pt} 
    \resizebox{0.85\textwidth}{!}{
    \begin{tabular}{l c c c }
    \toprule
        \textbf{Method} & $\text{IoU}_I$ & $\text{IoU}_T$ & mIoU \\
    \midrule
        SAM2-ZeroShot (ICLR'25)~\cite{ravisam} & 59.2$\pm$0.7 & 30.5$\pm$3.2 & 47.7$\pm$1.2 \\
        SAM2-Adapter (ICCV'23)~\cite{chen2023sam} & 72.8$\pm$1.2 & 50.3$\pm$2.2 & 63.8$\pm$1.0 \\
        SurgSAM2 (NeurIPS W'24)~\cite{liusurgical} & 74.2$\pm$1.0 & 53.1$\pm$2.6 & 65.8$\pm$1.5 \\
        \textbf{SurgMLLM (Ours)} & \textbf{87.8$\pm$1.4} & \textbf{67.8$\pm$3.2} & \textbf{84.4$\pm$1.2} \\
    \bottomrule
    \end{tabular}
    }
\end{table}
These results validate that explicit MLLM-based reasoning combined with \texttt{[SEG]}-guided grounding enables coherent, comprehensive, and spatially consistent surgical scene understanding. In contrast to fragmented pipelines that optimize phase recognition, triplet prediction, and segmentation independently, our unified framework jointly models their intrinsic correlations, leading to more consistent semantic predictions and more reliable dense grounding across tasks.
\noindent\textbf{Qualitative Results.} As shown in Fig.~\ref{fig:qualitative}, CurConMix~\cite{jeon2025curconmix} struggles to precisely capture fine-grained triplet relations, while SurgSAM2~\cite{liusurgical} can produce reasonable localizations but its masks are sometimes incomplete or inaccurate. Benefiting from unified reasoning and grounding, SurgMLLM correctly infers the workflow phase and both triplets, and its grounded instrument and target masks closely match the ground-truth, indicating improved semantic consistency and more reliable dense surgical scene understanding. More qualitative results and video demonstrations are provided in the supplementary video.
\begin{table}[h]
    \centering
    \caption{Ablation study on a subset of tasks within the framework and \texttt{[SEG]} prompt design. Experiments are conducted on Fold 1 of CholecT45-Scene dataset. ``-'' indicates the corresponding task is not applied.}
    \label{tab:ablation_seg}
    \setlength{\tabcolsep}{3pt}
    \resizebox{\textwidth}{!}{
    \begin{tabular}{c | l c c c}
        \toprule
            &  & Phase Accuracy & Triplet $\text{AP}_{IVT}$ & Grounding mIoU \\
        \midrule
            & \textbf{Ours} & \textbf{84.0} & \textbf{46.4} & \textbf{83.6} \\
        \midrule
            \multirow{3}{*}{Tasks} 
            & w/o triplet & 80.7 & - & 80.4 \\
            & w/o phase & - & 41.8 & 81.6 \\
            & w/o triplet-entity grounding & 82.6 & 40.1 & - \\
        \midrule
            \multirow{2}{*}{\texttt{[SEG]}}  & w/o residual fusion & 82.7 & 44.2 & 82.4 \\
            & w/o per-frame token & 81.5 & 43.6 & 81.1 \\
        \bottomrule
    \end{tabular}}
\end{table}
\noindent\textbf{Ablation Study.}
We conduct ablation studies on Fold 1 of the CholecT45-Scene dataset to examine the effects of joint multi-task learning and the proposed \texttt{[SEG]} prompt design (Tab.~\ref{tab:ablation_seg}). Removing supervision for any single task (\textit{w/o triplet}, \textit{w/o phase}, or \textit{w/o triplet-entity grounding}) consistently degrades the others, confirming that phase recognition, triplet reasoning, and grounding are tightly coupled and mutually beneficial under unified optimization. We further evaluate the \texttt{[SEG]} design. Disabling residual fusion (\textit{w/o residual fusion}) or sharing a single token across the clip (\textit{w/o per-frame token}) both reduce performance, demonstrating per-frame semantic tokens combined with temporal fusion are essential for consistent interaction modeling and reliable pixel-level grounding.
\section{Conclusion}
This work demonstrates that surgical scene understanding benefits \textit{fundamentally} from unifying temporal modeling, interaction reasoning, and spatial grounding within a single framework rather than treating them as isolated tasks. By coupling structured MLLM-based reasoning with pixel-level grounding, SurgMLLM improves overall surgical scene understanding. The introduced CholecT45-Scene dataset further enables joint evaluation of reasoning and grounding, supporting future research on more integrated surgical video analysis models. Despite these gains, SurgMLLM currently depends on dense triplet-entity mask supervision. Future work will explore weakly and semi-supervised grounding to reduce annotation dependence and better exploit unlabeled surgical videos.
\bibliographystyle{splncs04}
\bibliography{ref}

@inproceedings{gao2021trans,
  title={Trans-svnet: Accurate phase recognition from surgical videos via hybrid embedding aggregation transformer},
  author={Gao, Xiaojie and Jin, Yueming and Long, Yonghao and Dou, Qi and Heng, Pheng-Ann},
  booktitle={International Conference on Medical Image Computing and Computer-Assisted Intervention},
  year={2021}
}

@article{ding2022exploring,
  title={Exploring segment-level semantics for online phase recognition from surgical videos},
  author={Ding, Xinpeng and Li, Xiaomeng},
  journal={IEEE Transactions on Medical Imaging},
  year={2022}
}

@article{nwoye2022rendezvous,
  title={Rendezvous: Attention mechanisms for the recognition of surgical action triplets in endoscopic videos},
  author={Nwoye, Chinedu Innocent and Yu, Tong and Gonzalez, Cristians and Seeliger, Barbara and Mascagni, Pietro and Mutter, Didier and Marescaux, Jacques and Padoy, Nicolas},
  journal={Medical Image Analysis},
  year={2022}
}

@article{gui2023mt4mtl,
  title={Mt4mtl-kd: A multi-teacher knowledge distillation framework for triplet recognition},
  author={Gui, Shuangchun and Wang, Zhenkun and Chen, Jixiang and Zhou, Xun and Zhang, Chen and Cao, Yi},
  journal={IEEE Transactions on Medical Imaging},
  year={2023}
}

@article{ceron2022real,
  title={Real-time instance segmentation of surgical instruments using attention and multi-scale feature fusion},
  author={Cer{\'o}n, Juan Carlos {\'A}ngeles and Ruiz, Gilberto Ochoa and Chang, Leonardo and Ali, Sharib},
  journal={Medical Image Analysis},
  year={2022}
}

@article{nwoye2023cholectriplet2021,
  title={CholecTriplet2021: A benchmark challenge for surgical action triplet recognition},
  author={Nwoye, Chinedu Innocent and Alapatt, Deepak and Yu, Tong and Vardazaryan, Armine and Xia, Fangfang and Zhao, Zixuan and Xia, Tong and Jia, Fucang and Yang, Yuxuan and Wang, Hao and others},
  journal={Medical Image Analysis},
  year={2023}
}

@inproceedings{sharma2023surgical,
  title={Surgical action triplet detection by mixed supervised learning of instrument-tissue interactions},
  author={Sharma, Saurav and Nwoye, Chinedu Innocent and Mutter, Didier and Padoy, Nicolas},
  booktitle={International Conference on Medical Image Computing and Computer-Assisted Intervention},
  year={2023}
}

@article{alabi2025grounding,
  title={Grounding Surgical Action Triplets with Instrument Instance Segmentation: A Dataset and Target-Aware Fusion Approach},
  author={Alabi, Oluwatosin and Wei, Meng and Budd, Charlie and Vercauteren, Tom and Shi, Miaojing},
  journal={arXiv preprint arXiv:2511.00643},
  year={2025}
}

@inproceedings{kirillov2023segment,
  title={Segment anything},
  author={Kirillov, Alexander and Mintun, Eric and Ravi, Nikhila and Mao, Hanzi and Rolland, Chloe and Gustafson, Laura and Xiao, Tete and Whitehead, Spencer and Berg, Alexander C and Lo, Wan-Yen and others},
  booktitle={Proceedings of the IEEE/CVF International Conference on Computer Vision},
  year={2023}
}

@inproceedings{ravisam,
  title={SAM 2: Segment Anything in Images and Videos},
  author={Ravi, Nikhila and Gabeur, Valentin and Hu, Yuan-Ting and Hu, Ronghang and Ryali, Chaitanya and Ma, Tengyu and Khedr, Haitham and R{\"a}dle, Roman and Rolland, Chloe and Gustafson, Laura and others},
  booktitle={The Thirteenth International Conference on Learning Representations},
  year={2025}
}

@inproceedings{rasheed2024glamm,
  title={Glamm: Pixel grounding large multimodal model},
  author={Rasheed, Hanoona and Maaz, Muhammad and Shaji, Sahal and Shaker, Abdelrahman and Khan, Salman and Cholakkal, Hisham and Anwer, Rao M and Xing, Eric and Yang, Ming-Hsuan and Khan, Fahad S},
  booktitle={Proceedings of the IEEE/CVF Conference on Computer Vision and Pattern Recognition},
  year={2024}
}

@article{liu2023visual,
  title={Visual instruction tuning},
  author={Liu, Haotian and Li, Chunyuan and Wu, Qingyang and Lee, Yong Jae},
  journal={Advances in Neural Information Processing Systems},
  year={2023}
}

@article{yuan2025sa2va,
  title={Sa2va: Marrying sam2 with llava for dense grounded understanding of images and videos},
  author={Yuan, Haobo and Li, Xiangtai and Zhang, Tao and Sun, Yueyi and Huang, Zilong and Xu, Shilin and Ji, Shunping and Tong, Yunhai and Qi, Lu and Feng, Jiashi and others},
  journal={arXiv preprint arXiv:2501.04001},
  year={2025}
}

@inproceedings{lai2024lisa,
  title={Lisa: Reasoning segmentation via large language model},
  author={Lai, Xin and Tian, Zhuotao and Chen, Yukang and Li, Yanwei and Yuan, Yuhui and Liu, Shu and Jia, Jiaya},
  booktitle={Proceedings of the IEEE/CVF Conference on Computer Vision and Pattern Recognition},
  year={2024}
}

@inproceedings{gui2024tail,
  title={Tail-Enhanced Representation Learning for Surgical Triplet Recognition},
  author={Gui, Shuangchun and Wang, Zhenkun},
  booktitle={International Conference on Medical Image Computing and Computer-Assisted Intervention},
  year={2024}
}

@article{lan2025new,
  title={A new dataset and versatile multi-task surgical workflow analysis framework for thoracoscopic mitral valvuloplasty},
  author={Lan, Meng and Si, Weixin and Yan, Xinjian and Li, Xiaomeng},
  journal={Medical Image Analysis},
  year={2025}
}

@inproceedings{jeon2025curconmix,
  title={CurConMix: A Curriculum Contrastive Learning Framework for Enhancing Surgical Action Triplet Recognition},
  author={Jeon, Yongjun and Shin, Jongmin and Park, Seonmin and Kim, Bogeun and Park, Kanggil and Oh, Namkee and Jung, Kyu-Hwan},
  booktitle={International Conference on Medical Image Computing and Computer-Assisted Intervention},
  year={2025}
}

@inproceedings{chen2023sam,
  title={Sam-adapter: Adapting segment anything in underperformed scenes},
  author={Chen, Tianrun and Zhu, Lanyun and Deng, Chaotao and Cao, Runlong and Wang, Yan and Zhang, Shangzhan and Li, Zejian and Sun, Lingyun and Zang, Ying and Mao, Papa},
  booktitle={Proceedings of the IEEE/CVF International Conference on Computer Vision},
  year={2023}
}

@inproceedings{liusurgical,
  title={Surgical SAM 2: Real-time Segment Anything in Surgical Video by Efficient Frame Pruning},
  author={Liu, Haofeng and Zhang, Erli and Wu, Junde and Hong, Mingxuan and Jin, Yueming},
  booktitle={Conference on Neural Information Processing Systems Workshop},
  year={2024}
}

@inproceedings{liuunipixel,
  title={UniPixel: Unified Object Referring and Segmentation for Pixel-Level Visual Reasoning},
  author={Liu, Ye and Ma, Zongyang and Pu, Junfu and Qi, Zhongang and Wu, Yang and Shan, Ying and Chen, Chang Wen},
  booktitle={The Thirty-ninth Annual Conference on Neural Information Processing Systems},
  year={2025}
}

@article{long2025surgical,
  title={Surgical embodied intelligence for generalized task autonomy in laparoscopic robot-assisted surgery},
  author={Long, Yonghao and Lin, Anran and Kwok, Derek Hang Chun and Zhang, Lin and Yang, Zhenya and Shi, Kejian and Song, Lei and Fu, Jiawei and Lin, Hongbin and Wei, Wang and others},
  journal={Science Robotics},
  year={2025}
}

@article{carstens2025artificial,
  title={Artificial intelligence for surgical scene understanding: a systematic review and reporting quality meta-analysis},
  author={Carstens, Matthias and Vasisht, Shubha and Zhang, Zheyuan and Barbur, Iulia and Reinke, Annika and Maier-Hein, Lena and Hashimoto, Daniel A and Kolbinger, Fiona R},
  journal={npj Digital Medicine},
  year={2025}
}

@article{chen2024expanding,
  title={Expanding performance boundaries of open-source multimodal models with model, data, and test-time scaling},
  author={Chen, Zhe and Wang, Weiyun and Cao, Yue and Liu, Yangzhou and Gao, Zhangwei and Cui, Erfei and Zhu, Jinguo and Ye, Shenglong and Tian, Hao and Liu, Zhaoyang and others},
  journal={arXiv preprint arXiv:2412.05271},
  year={2024}
}

@article{bai2025qwen3,
  title={Qwen3-vl technical report},
  author={Bai, Shuai and Cai, Yuxuan and Chen, Ruizhe and Chen, Keqin and Chen, Xionghui and Cheng, Zesen and Deng, Lianghao and Ding, Wei and Gao, Chang and Ge, Chunjiang and others},
  journal={arXiv preprint arXiv:2511.21631},
  year={2025}
}

@article{pan2026surgical,
  title={Surgical Data Science in Time-Critical Contexts: A Roadmap Toward Brain-Inspired Computing},
  author={Pan, Yi and Zou, Shi-Hao and Yang, Jia-Wen and Si, Wei-Xin and Zheng, Wei-Min},
  journal={Journal of Computer Science and Technology},
  year={2026}
}

@article{li2023structural,
  title={Structural and pixel relation modeling for semisupervised instrument segmentation from surgical videos},
  author={Li, Caizi and Li, Yaoqian and Liu, Ruiqiang and Wang, Guangsuo and Lv, Jianping and Jin, Yueming and Si, Weixin and Heng, Pheng-Ann},
  journal={IEEE Transactions on Instrumentation and Measurement},
  year={2023}
}
\end{document}